\documentclass[runningheads]{llncs}

\usepackage{amsmath}

\usepackage{esvect}

\usepackage{multirow}
\usepackage[table,xcdraw]{xcolor}

\usepackage{listings}
\usepackage{xcolor}
\usepackage{fancyvrb}
\usepackage{caption}
\usepackage{float}
\usepackage{makecell}

\newcommand{\task}[1]{\textcolor{violet}{#1}}

\usepackage[T1]{fontenc}
\usepackage{graphicx}
\begin{document}
\title{FIRESPARQL: A LLM-based Framework for SPARQL Query Generation over Scholarly Knowledge Graphs}
%
%

\authorrunning{X. Pan et al.}
\author{
Xueli Pan\inst{1}\orcidID{0000-0002-3736-7047} \and
Victor de Boer\inst{1}\orcidID{0000-0001-9079-039X} \and
Jacco van Ossenbruggen\inst{1}\orcidID{0000-0002-7748-4715}
}
\institute{Vrije Universiteit Amsterdam, De Boelelaan 1105, 1081 HV Amsterdam, Netherlands
\email{\{x.pan2, v.de.boer, jacco.van.ossenbruggen\}@vu.nl}}

\maketitle              
\begin{abstract}
Question answering over Scholarly Knowledge Graphs (SKGs) remains a challenging task due to the complexity of scholarly content and the intricate structure of these graphs. Large Language Model (LLM) approaches could be used to translate natural language questions (NLQs) into SPARQL queries; however, these LLM-based approaches struggle with SPARQL query generation due to limited exposure to SKG-specific content and the underlying schema. We identified two main types of errors in the LLM-generated SPARQL queries: (i) structural inconsistencies, such as missing or redundant triples in the queries, and (ii) semantic inaccuracies, where incorrect entities or properties are shown in the queries despite a correct query structure. To address these issues, we propose FIRESPARQL, a modular framework that supports fine-tuned LLMs as a core component, with optional context provided via retrieval-augmented generation (RAG) and a SPARQL query correction layer. We evaluate the framework on the SciQA Benchmark using various configurations (zero-shot, zero-shot with RAG, one-shot, fine-tuning, and fine-tuning with RAG) and compare the performance with baseline and state-of-the-art approaches. We measure query accuracy using BLEU and ROUGE metrics, and query result accuracy using relaxed exact match(RelaxedEM), with respect to the gold standards containing the NLQs, SPARQL queries, and the results of the queries. Experimental results demonstrate that fine-tuning achieves the highest overall performance, reaching 0.90 ROUGE-L for query accuracy and 0.85 RelaxedEM for result accuracy on the test set.

\end{abstract}

\section{Introduction}
\label{introductin}
Question Answering (QA) over Knowledge Graphs (KGs), which allows users to query structured data using natural language, has gained considerable attention \cite{huang2019knowledge,omar2023chatgpt}. 
The task of QA over KGs usually takes a natural language question (NLQ) as an input and translates it into formal queries—typically SPARQL—that retrieve precise answers from the underlying KG. 
Previous studies in this domain have been centered around large-scale and encyclopedic KGs such as DBpedia, Freebase, and Wikidata. 
In these generic KGs, QA systems benefit from extensive community resources, well-documented schema, and relatively simple entity-relation structures. 
Recently, the emergence of large language models (LLMs) has inspired a growing body of research exploring their potential to address the task of QA over KGs \cite{wang2024knowledge,omar2023chatgpt} and benchmarked on datasets such as LC-QuAD \cite{trivedi2017lc}, QALD \cite{perevalov2022qald}, and WebQuestions \cite{berant2013semantic}.

However, applying these techniques to QA to Scholarly Knowledge Graphs (SKGs) presents significant challenges due to the intricate nature of scholarly data and the complex structure of SKGs \cite{jiang2023structure,pliukhin2023improving,taffa2023leveraging}.
Unlike encyclopedic KGs, SKGs capture domain-specific, technical content—such as research contributions, research problems, methodologies, datasets, and evaluation—often represented in complex ontological structures.
Several studies have investigated the potential of using LLMs for this task, exploring optimization techniques such as zero-shot learning, few-shot learning, and fine-tuning \cite{taffa2023leveraging,Lehmann2024large}.
Despite the improvements of LLMs on QA tasks over SKGs, LLMs face limitations when handling KG-specific parsing due to their lack of direct access to entities within the knowledge graph and insufficient understanding of the ontological schema, particularly for low-resource SKGs like the Open Research Knowledge Graph (ORKG)\cite{Auer2023TheSS}.

Insights from our pilot experiment revealed two major categories of errors LLMs tend to make in this task: (i) Structural inconsistencies, where generated SPARQL queries contain missing or redundant triples, and (ii) Semantic inaccuracies, where queries reference incorrect entities or properties, despite following the correct structural form.
To address these limitations, we propose FIRESPARQL, a LLM-based modular framework for SPARQL query generation over SKGs. 
At its core, FIRESPARQL supports FIne-tuned LLMs adapted to the SKG domain and offers relevant context provided via REtrieval-augmented generation (RAG) and a lightweight SPARQL correction layer. 
These components are designed to improve both the structural and semantic accuracy of the generated queries.

We investigate the effectiveness of this framework using the SciQA Benchmark \cite{Auer2023TheSS}, comparing multiple configurations—including zero-shot, one-shot, and fine-tuned models with and without RAG—against baseline and state-of-the-art methods. 
We assess performance based on BLEU and ROUGE scores for SPARQL query accuracy, and use a relaxed Exact Match metric to evaluate the accuracy of the returned query results.
Our findings demonstrate that domain-specific fine-tuning yields the most consistent and robust performance, significantly enhancing both query accuracy and result accuracy. 
Notably, the best-performance configuration is fine-tuned LLaMA3-8B-Instruct with 15 training epochs, achieving 0.77, 0.91, 0.86, 0.90, and 0.85 on BLEU-4, ROUGE-1, ROUGE-2, ROUGE-L, and RelaxedEM(all), respectively.
However, our experiments reveal that incorporating RAG into either the zero-shot or fine-tuned model does not yield further improvements and can even degrade performance.

The main contributions of this paper are three-fold:
(1) We identify and systematically categorize the common error types in LLM-generated SPARQL queries for QA over SKGs, distinguishing between structural inconsistencies and semantic inaccuracies. 
(2) We propose FIRESPARQL, a modular framework for SPARQL query generation that integrates a core fine-tuned LLM with an optional RAG module and a lightweight SPARQL correction layer.
(3) We conduct comprehensive experiments on the SciQA Benchmark under multiple configurations—including zero-shot, one-shot, fine-tuning, and their RAG-augmented variants—benchmarking against baselines and state-of-the-art methods using different model sizes and training epochs.

All resources and codes are available in our GitHub repository \footnote{\url{https://anonymous.4open.science/r/FIRESPARQL-7588}}.
For reproducibility, we have released the best-performing fine-tuned model—LLaMA-3-8B-Instruct trained for 15 epochs—on Hugging Face.

\section{Related Work}
\subsection{Traditional methods for QA over KGs}
Before the emergence of LLMs, QA over KGs is primarily addressed through knowledge graph embedding(KGE), neural network modeling, and reinforcement learning (RL).
These methods typically relied on modeling the structure of the KG and carefully engineered the features of the KG for entity linking, relation prediction, and path ranking.  
KGE-based approaches transform entities and relations into low-dimensional vector spaces to support efficient reasoning. 
Huang et al. \cite{huang2019knowledge} propose the KEQA framework for answering the most common types of questions by jointly recovering the question's head entity, predicate, and tail entity representations in the KG embedding spaces.
Graph neural networks (GNNs) have also been leveraged to reason over the structure of KGs.
Yasunaga et al. \cite{yasunaga2021qa} introduce QA-GNN, leveraging joint reasoning, where the QA context and KG are connected to form a joint graph and mutually update their representation through GNNs. 
Some studies emphasized RL to navigate the KG and identify answer paths.
Hai et al. \cite{cui2023path} propose AR2N, an interpretable reasoning method based on adversarial RL for multi-hop KGQA, to address the issue of spurious paths. AR2N consists of an answer generator and a path discriminator, which could effectively distinguish whether the reasoning chain is correct or not.

\subsection{QA over generic KGs such as DBpedia and Wikidata}
QA over generic Knowledge Graphs (KGs), such as DBpedia and Wikidata, has been extensively studied and serves as the foundation for many advances in the field. 
Early systems primarily focused on semantic parsing and graph-based reasoning [2, 14]. 
More recently, attention has shifted to neural and LLM-based approaches. Bustamante and Takeda [3] explore the use of entity-aware pre-trained GPT models for SPARQL generation, showing notable improvements in handling KG-specific structures. 
Similarly, Hovcevar and Kenda [4] integrate LLMs with KGs in industrial settings to support natural language interfaces. 
Meyer et al. [9] assess the SPARQL generation capabilities of various LLMs, pointing out both strengths and limitations in terms of structural correctness and execution accuracy. 
Other studies such as Kakalis and Kefalidis [7] focus on domain-specific extensions, like GeoSPARQL, and propose techniques for automatic URI injection. 
All in all, these works contribute a rich landscape of methods for mapping natural language questions to structured SPARQL queries across diverse knowledge bases.

\subsection{SPARQL generation for QA over SKGs}
QA over SKGs, such as the Open Research Knowledge Graph (ORKG), has gained attention due to its potential to support scientific exploration and knowledge discovery. 
Unlike generic knowledge graphs, SKGs capture fine-grained scholarly information, which introduces additional complexity in terms of schema diversity and domain-specific terminology. 
With the advent of LLMs, there has been a surge of interest in applying these models on the task of QA over SKGs. 
Lehmann et al. \cite{Lehmann2024large} conduct a comprehensive evaluation on the effectiveness of LLMs on the SciQA benchmark, demonstrating the models’ strengths in generating fluent SPARQL queries but also noting common pitfalls such as entity disambiguation and schema misalignment. 
Taffa and Usbeck \cite{taffa2023leveraging} specifically focus on adapting LLMs to the scholarly setting, emphasizing the need for domain-specific prompts and training data. 
Meanwhile, Pliukhin et al. \cite{pliukhin2023improving} explore fine-tuning strategies that improve LLM performance in one-shot scenarios. 
These studies collectively suggest that while LLMs offer promising capabilities, their effectiveness in the scholarly domain hinges on adaptation through fine-tuning, prompt engineering, and schema-aware correction mechanisms.

\section{Error type analysis on generated SPARQL queries}
\label{pilot-study}
Despite the improvements of LLMs on QA over SKGs, LLMs face limitations when handling KG-specific parsing.
The experimental results conducted by Sören Auer et al.\cite{Auer2023TheSS} showed that only 63 out of 100 handcrafted questions could be answered by ChatGPT, of which only 14 answers were correct.
To better understand why LLMs fail to generate the correct SPARQL query to a NLQ, we conduct a pilot experiment on using ChatGPT(GPT-4) with a random one-shot example to generate SPARQL queries for 30 handcrafted questions in the SciQA benchmark datasets.

Insights from this pilot experiment revealed two major categories of errors LLMs tend to make in this task: semantic inaccuracies and structural inconsistencies.
Semantic inaccuracies occur when LLMs fail to link the correct properties and entities in ORKG, despite generating SPARQL queries with correct structure.
Our observations reveal that LLMs tend to rely on the example provided in the one-shot learning process to generate the correct structure for a certain type of questions, but often struggle with linking the correct properties and entities because LLMs do not learn the content of the underlying KG.
Structural inconsistencies arise due to LLMs’ lack of ontological schema of the underlying KG, leading to errors in query structure, such as missing or abundant links (triples), despite correctly linking to the mentioned entities or properties.

\begin{figure}
    \centering
    \includegraphics[width=1.1\linewidth]{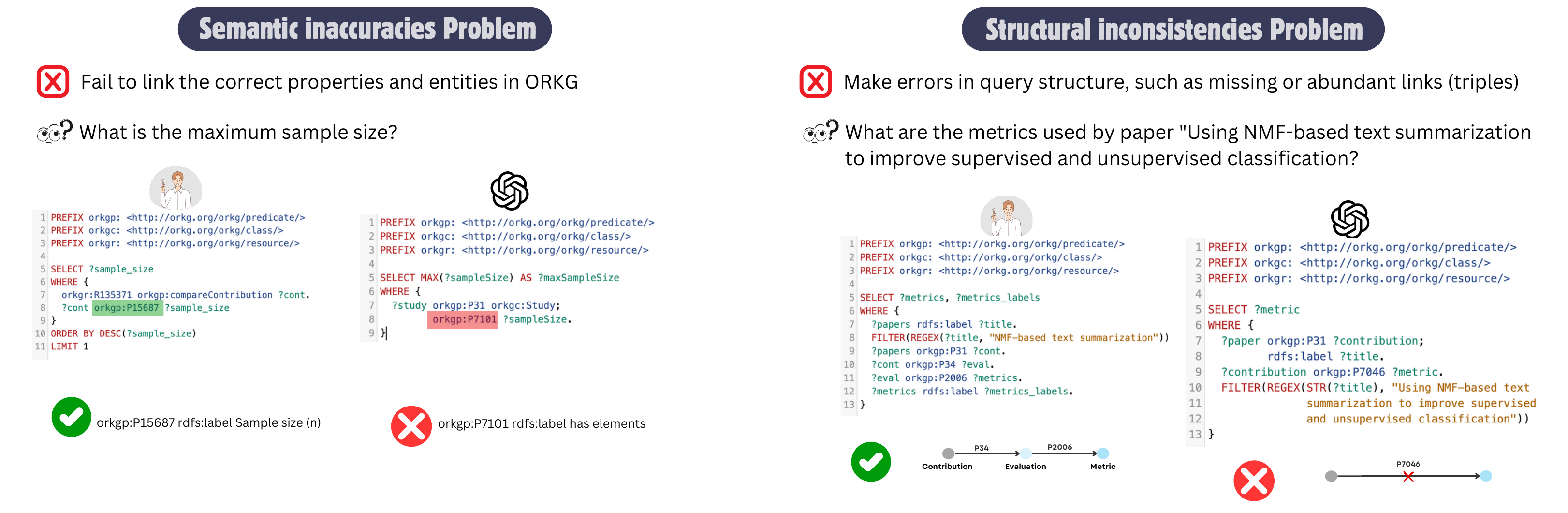}
    \caption{Examples of semantic inaccuracies and structural inconsistencies problem with the generted SPARQL queries}
    \label{fig:error-types}
\end{figure}

Figure \ref{fig:error-types} shows the example of semantic inaccuracies and structural inconsistencies problem with the generated SPARQL queries in our pilot study.
In the example of the semantic inaccuracies problem, ChatGPT failed to link the correct property orkgp:P15687; instead, it linked to a wrong property orkgp:P7101.
In the example of the structural inconsistencies problem, the SPARQL query generated by ChatGPT directly links Contribution to Metrics, fails to detect the correct schema of the ORKG where Contribution and Metric are connected via Evaluation.

\section{Methodology}
\begin{figure}
    \centering
    \includegraphics[width=1.2\linewidth]{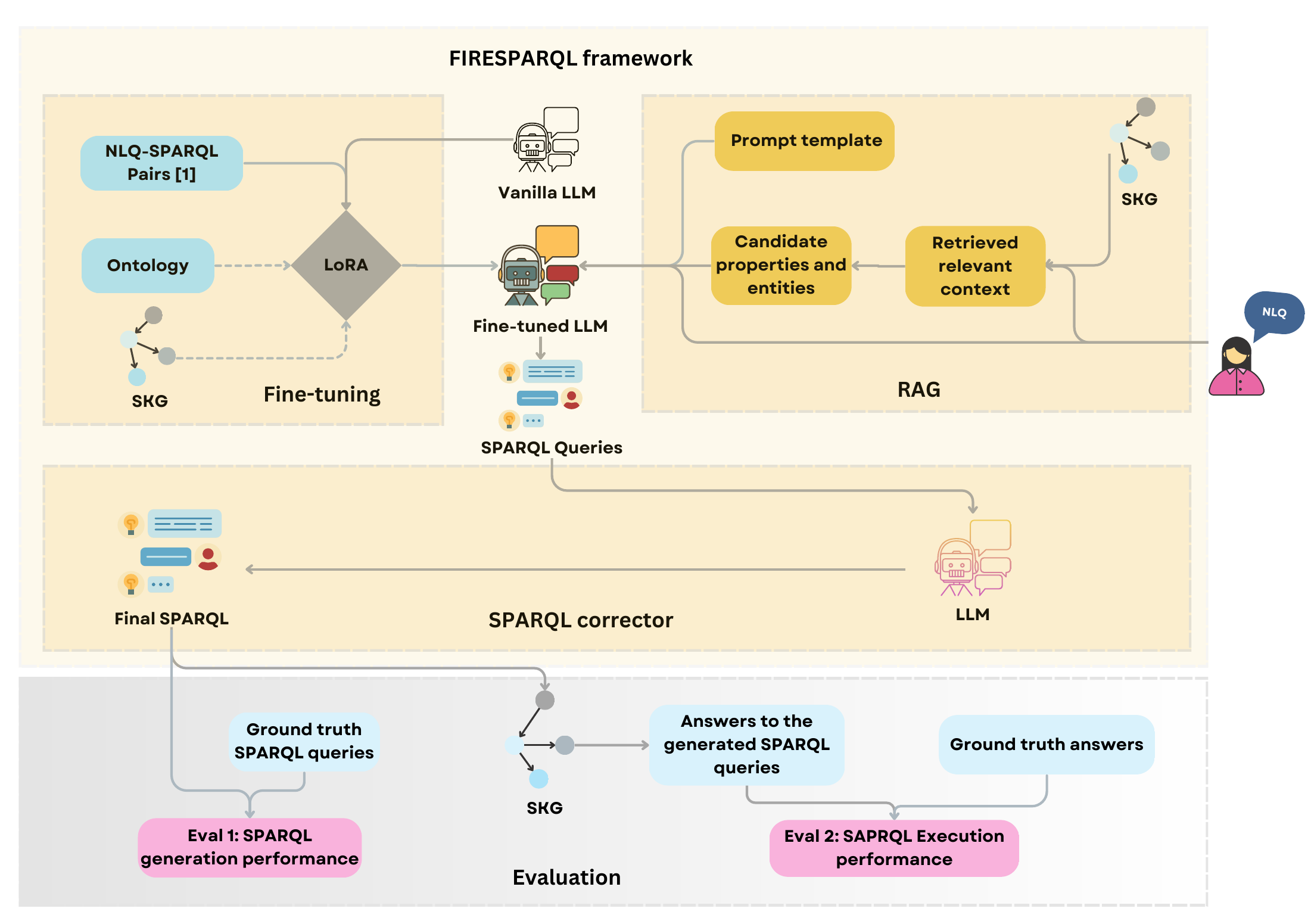}
    \caption{FIRESPARQL framework (yellow boxes) and evaluation setup (grey boxes)}
    \label{fig:firesparql}
\end{figure}

As we mentioned in Section \ref{pilot-study}, generating executable and semantically accurate SPARQL queries over SKGs remains a challenging task due to two main types of errors: semantic inaccuracies and structural inconsistencies.
To address these issues, we propose FIRESPARQL, a modular framework designed to improve both the semantic accuracy and the structural consistency of generated SPARQL queries. 
The framework consists of three core components: (1) Fine-tuned LLMs, (2) Retrieval-Augmented Generation (RAG), and (3) SPARQL correction. 
The final SPARQL queries are evaluated at both the query accuracy and execution accuracy using ground truth comparisons. 
An overview of the framework and the evaluation setup is shown in Figure \ref{fig:firesparql}.

\subsection{Fine-tuning}
At the core of FIRESPARQL is a fine-tuning module applying Low-Rank Adaptation (LoRA) \cite{hu2021lora} for parameter-efficient fine-tuning.
Unlike full-parameter fine-tuning, LoRA freezes the pre-trained model weights and injects trainable low-rank decomposition matrices into each layer of the Transformer architecture. 
This approach significantly reduces the number of trainable parameters required for downstream tasks while maintaining strong performance.
Fine-tuning LLMs has proven effective in scientific knowledge extraction \cite{muralidharan2024knowledgeaifinetuningnlp} and KG construction \cite{ghanem:hal-04862235,wang2024techgpt20largelanguagemodel}.

To address structural inconsistencies in generated SPARQL queries—often arising from a limited understanding of the SKG schema—we fine-tune LLMs so that the ontology and structural patterns of the underlying SKG are implicitly captured during training.
The fine-tuning data can include the ontology descriptions, RDF triples, or task-specific labeled examples. 
In our implementation, we use NLQ-SPARQL query pairs as training data, which implicitly encode the structure and vocabulary of the target SKG. This results in a fine-tuned LLM capable of generating syntactically correct and semantically meaningful SPARQL queries directly from natural language questions.

We further investigate the impact of training epochs on fine-tuning performance. 
The number of epochs determines how many times the model iterates over the training data, directly influencing its ability to capture domain-specific patterns.
The prompt template for SPARQL generation is shown in Listing \ref{lst:prompt-sparql}.

\vspace{10pt}
\begin{lstlisting}[
%basicstyle=\footnotesize, 
caption={Prompt template for SPARQL generation}, label={lst:prompt-sparql}]
The Open Research Knowledge Graph (ORKG) is a semantic knowledge graph designed to represent, compare, and retrieve scholarly contributions. Given a natural language question in English, your task is to generate the corresponding SPARQL query to this question. The generated SPARQL query should be able to query the ORKG, getting correct answer to the input question. 
Give me only the SPARQL query, no other text.
Input question: {(*@\task{input question}@*)}
Output SPARQL query:
\end{lstlisting}

\subsection{RAG}
RAG \cite{lewis2020retrieval} has been proposed to enable LLMs access to external and domain-specific knowledge for knowledge-intensive NLP tasks, which could be a promising way to address the issue of semantic inaccuracies—where generated SPARQL queries fail to link to the correct properties or entities. 
These inaccuracies often stem from the model’s limited exposure to SKGs or ambiguous entity/property mentions in the input question.
Therefore, we propose an optional RAG module in the framework to enhance the model’s contextual understanding of the underlying SKGs.
Given an NLQ, relevant context is retrieved from the SKG in the form of candidate entities, properties, or subgraphs. 
A prompt template then incorporates this contextual information alongside the input question, which is passed to the finetuned LLM to generate a more semantically accurate SPARQL query.
In our implementation, we use RAG to retrieve candidate properties from a curated list of distinct ORKG properties, including their URLs and labels. 
These candidates are then incorporated as contextual information in the prompt template to guide SPARQL generation.

\subsection{SPARQL cleaning}
Despite improvements through fine-tuning and RAG, generated SPARQL queries may still contain minor structural or syntactic errors that hinder successful execution. 
These include unnecessary text in the output, missing or extra punctuation, or subtle syntax issues such as missing spaces between variable names.
To address this, we introduce a lightweight SPARQL correction layer based on LLMs. 
This module takes the initially generated query as input and refines it to ensure syntactic validity, which increases the likelihood of generating executable SPARQL queries. 
The cleaned queries are then passed to the evaluation stage.

\section{Experiments}
\subsection{Datasets}
We conduct experiments on the SciQA benchmark dataset, a recently released resource designed to evaluate question answering systems over scholarly knowledge graphs \cite{Auer2023TheSS}.
SciQA provides a diverse set of natural language questions aligned with the Open Research Knowledge Graph (ORKG).
SciQA contains 100 handcrafted natural language questions with paraphrases, corresponding human- and machine-readable SPARQL queries with their results.
In addition, a set of 2465 questions has been semi-automatically derived from eight question templates. 

\subsection{Baselines and the state of the art}
As a baseline, we adopt a zero-shot setting in which the model generates SPARQL queries without any task-specific fine-tuning or example guidance. This setup evaluates the model's out-of-the-box ability to understand the task and map natural language questions to structured SPARQL queries. 
For the state-of-the-art comparison, we implement a one-shot approach in which the model is provided with the most semantically similar question from the training set, along with its corresponding SPARQL query, as an in-context demonstration. 
The most similar example is identified by computing cosine similarity between the input question and all training questions using SentenceBERT embeddings.
This configuration has shown strong performance in recent studies by helping the model better understand the tasks \cite{Lehmann2024large,liu2022makes}.

\subsection{Implementation}
We fine-tuned two instruction-tuned models, Llama 3.2-3B Instruct and Llama 3-8B Instruct, using 1,795 labeled samples from the SciQA training set. The models were trained under various epoch configurations (3, 5, 7, 10, 15, and 20) to analyze performance across training durations. 
All fine-tuning experiments were conducted on a single NVIDIA H100 GPU.
We used DeepSeek-R1-Distill-Llama-70B as the underlying model for retrieval-augmented generation (RAG).
\textbf{All SPARQL queries are executed via Qlever \cite{bast2017qlever}}

\subsection{Evaluation metrics}
We employ a combination of string-based and execution-based evaluation metrics to assess the quality of the generated SPARQL queries. 
Similar to other research, we use BLEU-4 and ROUGE scores, which measure token-level and n-gram overlaps, to evaluate the similarity between generated queries and the ground-truth queries.
These metrics provide insights into how closely the structure and content of the generated queries align with the reference queries. 
Additionally, we assess the execution performance of the generated SPARQL queries using two variants of relaxed Exact Match: Success and All. 
The Relaxed Exact Match (success) metric considers only those queries that were syntactically valid, successfully executed against the ORKG RDF dump, and returned non-empty results. 
In contrast, the Relaxed Exact Match (all) metric evaluates the correctness of the query results across the entire test set, including queries that may have failed or returned empty results. 

Unlike the original Exact Match, which is very strict, our Relaxed Exact Match incorporates several preprocessing steps. 
First, we remove variable names from the returned results to avoid penalizing differences that do not affect semantics. 
Second, we split the results line by line and eliminate duplicate lines to normalize the output structure. 
Finally, we compare the processed sets using exact matching. 
This approach provides a more tolerant and realistic evaluation of query execution performance in scholarly knowledge graph settings.
The above-mentioned dual evaluation approach allows us to comprehensively analyze both the syntactic quality and the practical effectiveness of the generated SPARQL queries.

\subsection{Results}
Table \ref{table:bleu-rouge-em} shows the results of different metrics on different strategies, different models, and different epochs.
Figure \ref{fig:bleu-rouge-em} shows the results of different metrics with different epochs for LoRA fine-tuning.
All the scores are the average scores of three runs.
The standard deviation of BLEU-4, ROUGE scores, RelaxedEM(success), and RelaxedEM(all) across different model variants (e.g., zero-shot, one-shot, fine-tuning, and RAG) and training epochs is consistently low ($std < 0.0265$), indicating stable performance across all three runs. 
Therefore, we use the average scores as a reliable and representative summary of model effectiveness. In the next section, we discuss the main takeaways from these results.

\begin{table}[]
\tiny
\begin{tabular}{|
>{\columncolor[HTML]{FFFFFF}}c |
>{\columncolor[HTML]{FFFFFF}}c |
>{\columncolor[HTML]{FFFFFF}}c |
>{\columncolor[HTML]{FFFFFF}}c |
>{\columncolor[HTML]{FFFFFF}}c |
>{\columncolor[HTML]{FFFFFF}}c |
>{\columncolor[HTML]{FFFFFF}}c |
>{\columncolor[HTML]{FFFFFF}}c |
>{\columncolor[HTML]{FFFFFF}}c |}
\hline
{\color[HTML]{000000} Strategy} & {\color[HTML]{000000} Model} & {\color[HTML]{000000} Epoch} & {\color[HTML]{000000} BLEU-4} & {\color[HTML]{000000} ROUGE-1} & {\color[HTML]{000000} ROUGE-2} & {\color[HTML]{000000} ROUGE-L} & {\color[HTML]{000000} \begin{tabular}[c]{@{}c@{}}RelaxedEM \\ (success)\end{tabular}} & {\color[HTML]{000000} \begin{tabular}[c]{@{}c@{}}RelaxedEM\\  (all)\end{tabular}} \\ \hline
\cellcolor[HTML]{FFFFFF}{\color[HTML]{000000} } & {\color[HTML]{000000} llama-3.2-3b-Instruct} & {\color[HTML]{000000} -} & {\color[HTML]{000000} 0.03} & {\color[HTML]{000000} 0.36} & {\color[HTML]{000000} 0.18} & {\color[HTML]{000000} 0.35} & {\color[HTML]{000000} 0.00} & {\color[HTML]{000000} 0.00} \\ \cline{2-9} 
\multirow{-2}{*}{\cellcolor[HTML]{FFFFFF}{\color[HTML]{000000} zero-shot}} & {\color[HTML]{000000} llama-3-8b-Instruct} & {\color[HTML]{000000} -} & {\color[HTML]{000000} 0.03} & {\color[HTML]{000000} 0.39} & {\color[HTML]{000000} 0.18} & {\color[HTML]{000000} 0.38} & {\color[HTML]{000000} 0.00} & {\color[HTML]{000000} 0.00} \\ \hline
\cellcolor[HTML]{FFFFFF}{\color[HTML]{000000} } & {\color[HTML]{000000} llama-3.2-3b-Instruct} & {\color[HTML]{000000} -} & {\color[HTML]{000000} 0.03} & {\color[HTML]{000000} 0.36} & {\color[HTML]{000000} 0.18} & {\color[HTML]{000000} 0.36} & {\color[HTML]{000000} 0.00} & {\color[HTML]{000000} 0.00} \\ \cline{2-9} 
\multirow{-2}{*}{\cellcolor[HTML]{FFFFFF}{\color[HTML]{000000} zero-shot\_rag}} & {\color[HTML]{000000} llama-3-8b-Instruct} & {\color[HTML]{000000} -} & {\color[HTML]{000000} 0.03} & {\color[HTML]{000000} 0.36} & {\color[HTML]{000000} 0.18} & {\color[HTML]{000000} 0.38} & {\color[HTML]{000000} 0.00} & {\color[HTML]{000000} 0.00} \\ \hline
\cellcolor[HTML]{FFFFFF}{\color[HTML]{000000} } & {\color[HTML]{000000} llama-3.2-3b-Instruct} & {\color[HTML]{000000} -} & {\color[HTML]{000000} 0.58} & {\color[HTML]{000000} 0.81} & {\color[HTML]{000000} 0.73} & {\color[HTML]{000000} 0.78} & {\color[HTML]{000000} 0.78} & {\color[HTML]{000000} 0.40} \\ \cline{2-9} 
\multirow{-2}{*}{\cellcolor[HTML]{FFFFFF}{\color[HTML]{000000} oneshot}} & {\color[HTML]{000000} llama-3-8b-Instruct} & {\color[HTML]{000000} -} & {\color[HTML]{000000} 0.38} & {\color[HTML]{000000} 0.61} & {\color[HTML]{000000} 0.50} & {\color[HTML]{000000} 0.59} & {\color[HTML]{000000} 0.89} & {\color[HTML]{000000} 0.29} \\ \hline
\cellcolor[HTML]{FFFFFF}{\color[HTML]{000000} } & {\color[HTML]{000000} \begin{tabular}[c]{@{}c@{}}llama-3.2-3b-Instruct-lora\_\\ deepseekr1-distill-llama-70b\end{tabular}} & {\color[HTML]{000000} 20} & {\color[HTML]{000000} 0.32} & {\color[HTML]{000000} 0.63} & {\color[HTML]{000000} 0.51} & {\color[HTML]{000000} 0.60} & {\color[HTML]{000000} 0.42} & {\color[HTML]{000000} 0.06} \\ \cline{2-9} 
\multirow{-2}{*}{\cellcolor[HTML]{FFFFFF}{\color[HTML]{000000} ft\_rag}} & {\color[HTML]{000000} \begin{tabular}[c]{@{}c@{}}llama3-8b-Instruct-lora\_\\ deepseekr1-disttill-llama-70b\end{tabular}} & {\color[HTML]{000000} 15} & {\color[HTML]{000000} 0.58} & {\color[HTML]{000000} 0.81} & {\color[HTML]{000000} 0.72} & {\color[HTML]{000000} 0.77} & {\color[HTML]{000000} 0.85} & {\color[HTML]{000000} 0.29} \\ \hline
\cellcolor[HTML]{FFFFFF}{\color[HTML]{000000} } & \cellcolor[HTML]{FFFFFF}{\color[HTML]{000000} } & {\color[HTML]{000000} 3} & {\color[HTML]{000000} 0.67} & {\color[HTML]{000000} 0.86} & {\color[HTML]{000000} 0.79} & {\color[HTML]{000000} 0.83} & {\color[HTML]{000000} 0.88} & {\color[HTML]{000000} 0.53} \\ \cline{3-9} 
\cellcolor[HTML]{FFFFFF}{\color[HTML]{000000} } & \cellcolor[HTML]{FFFFFF}{\color[HTML]{000000} } & {\color[HTML]{000000} 5} & {\color[HTML]{000000} 0.62} & {\color[HTML]{000000} 0.83} & {\color[HTML]{000000} 0.75} & {\color[HTML]{000000} 0.79} & {\color[HTML]{000000} 0.71} & {\color[HTML]{000000} 0.36} \\ \cline{3-9} 
\cellcolor[HTML]{FFFFFF}{\color[HTML]{000000} } & \cellcolor[HTML]{FFFFFF}{\color[HTML]{000000} } & {\color[HTML]{000000} 7} & {\color[HTML]{000000} 0.52} & {\color[HTML]{000000} 0.77} & {\color[HTML]{000000} 0.67} & {\color[HTML]{000000} 0.73} & {\color[HTML]{000000} 0.79} & {\color[HTML]{000000} 0.27} \\ \cline{3-9} 
\cellcolor[HTML]{FFFFFF}{\color[HTML]{000000} } & \cellcolor[HTML]{FFFFFF}{\color[HTML]{000000} } & {\color[HTML]{000000} 10} & {\color[HTML]{000000} 0.56} & {\color[HTML]{000000} 0.79} & {\color[HTML]{000000} 0.70} & {\color[HTML]{000000} 0.76} & {\color[HTML]{000000} 0.70} & {\color[HTML]{000000} 0.22} \\ \cline{3-9} 
\cellcolor[HTML]{FFFFFF}{\color[HTML]{000000} } & \cellcolor[HTML]{FFFFFF}{\color[HTML]{000000} } & {\color[HTML]{000000} 15} & {\color[HTML]{000000} 0.60} & {\color[HTML]{000000} 0.81} & {\color[HTML]{000000} 0.73} & {\color[HTML]{000000} 0.78} & {\color[HTML]{000000} 0.88} & {\color[HTML]{000000} 0.40} \\ \cline{3-9} 
\cellcolor[HTML]{FFFFFF}{\color[HTML]{000000} } & \multirow{-6}{*}{\cellcolor[HTML]{FFFFFF}{\color[HTML]{000000} llama-3.2-3b-Instruct-lora}} & {\color[HTML]{000000} 20} & {\color[HTML]{000000} \textbf{0.70}} & {\color[HTML]{000000} \textbf{0.86}} & {\color[HTML]{000000} \textbf{0.79}} & {\color[HTML]{000000} \textbf{0.84}} & {\color[HTML]{000000} \textbf{0.80}} & {\color[HTML]{000000} \textbf{0.54}} \\ \cline{2-9} 
\cellcolor[HTML]{FFFFFF}{\color[HTML]{000000} } & \cellcolor[HTML]{FFFFFF}{\color[HTML]{000000} } & {\color[HTML]{000000} 3} & {\color[HTML]{000000} 0.75} & {\color[HTML]{000000} 0.90} & {\color[HTML]{000000} 0.84} & {\color[HTML]{000000} 0.88} & {\color[HTML]{000000} 0.99} & {\color[HTML]{000000} 0.79} \\ \cline{3-9} 
\cellcolor[HTML]{FFFFFF}{\color[HTML]{000000} } & \cellcolor[HTML]{FFFFFF}{\color[HTML]{000000} } & {\color[HTML]{000000} 5} & {\color[HTML]{000000} 0.74} & {\color[HTML]{000000} 0.90} & {\color[HTML]{000000} 0.85} & {\color[HTML]{000000} 0.87} & {\color[HTML]{000000} 0.98} & {\color[HTML]{000000} 0.73} \\ \cline{3-9} 
\cellcolor[HTML]{FFFFFF}{\color[HTML]{000000} } & \cellcolor[HTML]{FFFFFF}{\color[HTML]{000000} } & {\color[HTML]{000000} 7} & {\color[HTML]{000000} 0.70} & {\color[HTML]{000000} 0.87} & {\color[HTML]{000000} 0.80} & {\color[HTML]{000000} 0.84} & {\color[HTML]{000000} 0.96} & {\color[HTML]{000000} 0.69} \\ \cline{3-9} 
\cellcolor[HTML]{FFFFFF}{\color[HTML]{000000} } & \cellcolor[HTML]{FFFFFF}{\color[HTML]{000000} } & {\color[HTML]{000000} 10} & {\color[HTML]{000000} 0.71} & {\color[HTML]{000000} 0.88} & {\color[HTML]{000000} 0.81} & {\color[HTML]{000000} 0.85} & {\color[HTML]{000000} 0.94} & {\color[HTML]{000000} 0.69} \\ \cline{3-9} 
\cellcolor[HTML]{FFFFFF}{\color[HTML]{000000} } & \cellcolor[HTML]{FFFFFF}{\color[HTML]{000000} } & {\color[HTML]{000000} 15} & {\color[HTML]{000000} \textbf{0.77}} & {\color[HTML]{000000} \textbf{0.91}} & {\color[HTML]{000000} \textbf{0.86}} & {\color[HTML]{000000} \textbf{0.90}} & {\color[HTML]{000000} \textbf{0.98}} & {\color[HTML]{000000} \textbf{0.85}} \\ \cline{3-9} 
\multirow{-12}{*}{\cellcolor[HTML]{FFFFFF}{\color[HTML]{000000} ft}} & \multirow{-6}{*}{\cellcolor[HTML]{FFFFFF}{\color[HTML]{000000} llama-3-8b-Instruct-lora}} & {\color[HTML]{000000} 20} & {\color[HTML]{000000} 0.74} & {\color[HTML]{000000} 0.89} & {\color[HTML]{000000} 0.84} & {\color[HTML]{000000} 0.88} & {\color[HTML]{000000} 0.99} & {\color[HTML]{000000} 0.82} \\ \hline
\end{tabular}
\vspace{10pt}
\caption{BLEU, ROUGE and RelaxedEM scores with different strategies, different models and different epochs}
\label{table:bleu-rouge-em}
\end{table}

\section{Discussion}

We discuss the key findings from the experiments and discuss limitations and future work.

\subsection{Discussion on the results}
Our experimental findings provide several key insights into the effectiveness of the FIRESPARQL framework and the performance of different strategies for SPARQL query generation with different epoch settings and different model sizes.

\subsubsection{Fine-Tuning Performance}
As shown in Table \ref{table:bleu-rouge-em}, fine-tuning LLMs on NLQ–SPARQL pairs leads to significant improvements over both the zero-shot baseline and the one-shot state-of-the-art methods. 
The highest performance is achieved by the fine-tuned LLaMA-3-8B-Instruct model trained for 15 epochs, attaining scores of 0.77 (BLEU-4), 0.91 (ROUGE-1), 0.86 (ROUGE-2), 0.90 (ROUGE-L), and 0.85 (RelaxedEM on all test cases).
These results indicate that, across both syntactic-level (BLEU, ROUGE) and execution-level (RelaxedEM) evaluations, SPARQL queries generated by finetuned models are not only more accurate but also structurally well-formed and executable. 
This highlights the effectiveness of supervised adaptation for learning the ontology and structure of the underlying SKG during training.

\subsubsection{Model Size Impact}
\begin{figure}
    \centering
    \includegraphics[width=1.0\linewidth]{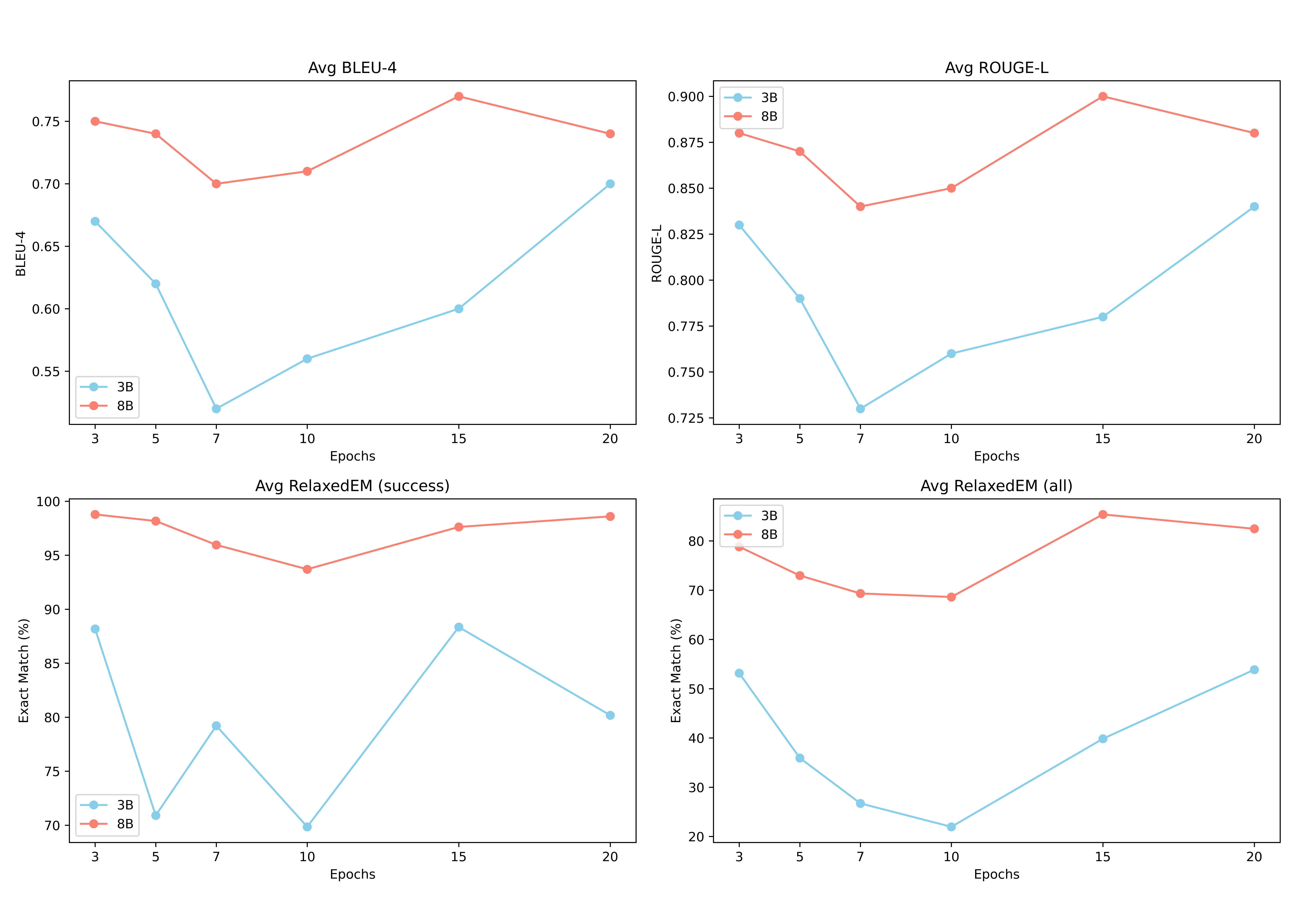}
    \caption{Average BLEU-4, ROUGE-L and RelaxedEM scores with different epochs on different fine-tuned models}
    \label{fig:bleu-rouge-em}
\end{figure}

As shown in Fig \ref{fig:bleu-rouge-em}, LLaMA-3-8B-Instruct consistently outperforms LLaMA-3.2-3B-Instruct after fine-tuning across all evaluation metrics. 
This demonstrates that larger model capacity enhances the ability to internalize domain-specific patterns from training data, including the structure and semantics of the target SKG.
Interestingly, the trend reverses in the one-shot setting: LLaMA-3.2-3B-Instruct performs better than the 8B variant on most metrics, except for RelaxedEM(success), as shown in Table \ref{table:bleu-rouge-em}.
This performance gap might be attributed to the fact that LLaMA-3.2-3B-Instruct was released after LLaMA-3-8B-Instruct and incorporates pruning and distillation techniques, which were specifically applied to the 1B and 3B variants \cite{llama}. 
These techniques help to preserve performance while significantly improving efficiency, making the LLaMA-3.2-3B-Instruct model capable of strong instruction-following performance on resource-constrained devices. As a result, despite its smaller size, LLaMA-3.2-3B-Instruct may benefit from a more refined architecture and training strategies, allowing it to better leverage in-context examples in one-shot settings compared to the larger, but earlier, 8B model.

\subsubsection{RAG Performance}
As shown in Table \ref{table:bleu-rouge-em}, the score of RelaxedEM(all) drops from 0.85 to 0.29 when incorporating RAG into the fine-tuned LLaMA-3-8B-Instruct trained for 15 epochs, which does not lead to additional performance gains, and it even degrades performance. 
This decline can be attributed to the noisy or misaligned nature of the retrieved context—such as incorrect or irrelevant property suggestions from the ORKG—which may introduce confusion instead of providing useful guidance since we don't have a context checker to check if the context is relevant or not.
Prior studies \cite{jin2024long,joren2024sufficient} have similarly observed that low-quality RAG context can conflict with the knowledge already encoded in fine-tuned models, ultimately leading to reduced task performance.

\subsubsection{One-Shot Performance}
As shown in Table \ref{table:bleu-rouge-em}, the one-shot setting—using the most similar example from the training set—achieved strong performance, second only to the fine-tuned models. 
Specifically, the one-shot approach reached scores of 0.58 (BLEU-4), 0.81 (ROUGE-1), 0.73 (ROUGE-2), 0.78 (ROUGE-L), and 0.40 (RelaxedEM on all test cases) on LLaMA-3.2-3B-Instruct model.
These results suggest that in the absence of high-quality datasets for fine-tuning, one-shot learning offers a simple yet effective alternative for SPARQL query generation.

\subsubsection{Training Epoch Sensitivity}
As shown in Figure \ref{fig:bleu-rouge-em}, the number of fine-tuning epochs has a significant impact on all metrics for both LLaMA-3.2-3B-Instruct and LLaMA-3-8B-Instruct models.
First, both models start with high scores on all metrics at epoch 3 and then slightly decline during the early training phases (epochs 3–7), suggesting that the models may require sufficient training time to properly internalize the SKG-specific structures and semantics.
Second, the training dynamics reveal an upward trend in performance from 7 epochs onward, with the best performance at 20 epochs for the 3B model and 15 epochs for the 8B model. 
This indicates that larger models tend to converge faster and exhibit stronger generalization early on, while smaller models require more epochs to achieve competitive performance.

\subsection{Error analysis on failed SPARQL queries}
We further analyzed the generated SPARQL queries that either failed to execute or returned empty results, by comparing them with the corresponding ground truth queries. Under the best-performing configuration—using the fine-tuned \texttt{LLaMA3-8B-Instruct} model trained for 15 epochs—448 out of 513 generated SPARQL queries were executed successfully via QLever without any syntax errors and returned meaningful results. Meanwhile, 14 queries failed due to syntax errors, and 51 queries were executed successfully but returned empty results.

To better understand the causes of failure, we examined the error messages for the 14 syntactically invalid queries and inspected the queries that returned empty results. Our analysis revealed that 11 out of the 14 syntactically invalid queries shared the same error message:
\begin{quote}
\small
\texttt{Invalid SPARQL query: Variable ?metric is selected but not aggregated. All non-aggregated variables must be part of the GROUP BY clause. Note: The GROUP BY in this query is implicit because an aggregate expression was used in the SELECT clause.}
\end{quote}
This indicates that these queries included aggregate functions (e.g., \texttt{MAX(?value)}) in the \texttt{SELECT} clause but did not include the non-aggregated variables (e.g., \texttt{?metric}, \texttt{?metric\_lbl}) in a \texttt{GROUP BY} clause. In SPARQL 1.1, such usage is invalid unless all non-aggregated variables are explicitly grouped. This reflects a lack of adherence to SPARQL's syntax rules around aggregation and grouping.

The remaining 3 queries failed with the following error:
\begin{quote}
\small
\texttt{Invalid SPARQL query: Token 'SELECT': mismatched input 'SELECT' expecting '\}'.}
\end{quote}
This indicates that the queries contained improperly structured subqueries.
Specifically, full \texttt{SELECT} statements nested directly inside a \texttt{WHERE} clause without being enclosed in curly braces (\texttt{\{\}}). SPARQL requires subqueries to be syntactically isolated within their own scope using curly braces. 
These errors likely stem from incorrect handling of nested query structures during generation.

These findings highlight the current limitations of fine-tuned LLMs in capturing the formal syntactic constraints of the SPARQL query language, particularly in scenarios involving nested subqueries and aggregation functions.
As a potential extension to our approach, prompt engineering techniques that include explicit syntax error examples or constraint reminders could be incorporated during SPARQL generation to encourage the model to produce syntactically valid SPARQL, especially for complex constructs like aggregation and subqueries.

\subsection{Limitations and Future Work}
While FIRESPARQL demonstrates strong performance in generating SPARQL queries over the ORKG, several limitations remain, which also highlight directions for future research: 
Our current experiments are limited to a single domain-specific benchmark, SciQA, which is built on top of the ORKG. To assess the generalizability of our approach, it is crucial to evaluate FIRESPARQL on a broader range of benchmarks across different domains and knowledge graphs. 
This would help determine whether the observed improvements are transferable or highly task-specific.
Although our framework includes an optional RAG module, its effectiveness is currently hindered by the quality of the retrieved context. 
In many cases, irrelevant or incorrect candidate properties are introduced, leading to performance degradation. 
Future work should focus on developing more accurate and semantically aware retrieval mechanisms that can provide high-quality, contextually relevant information—such as topological subgraphs or query templates—without introducing noise.
FIRESPARQL relies on supervised fine-tuning using NLQ–SPARQL pairs, which may not always be available. 
In future work, we aim to explore alternative data for fine-tuning LLMs such as synthetic training data or leveraging weak supervision from ontology or subgraphs.

\section{Conclusion}
In this paper, we introduced FIRESPARQL, a modular framework for SPARQL query generation over SKGs. 
By systematically analyzing common error types, structural inconsistencies, and semantic inaccuracies, we designed a three-module architecture comprising fine-tuned LLMs for SPARQL generation, optional RAG for providing relevant context, and a lightweight SPARQL correction layer. 
Our empirical evaluation on the SciQA benchmark demonstrates that domain-specific fine-tuning, especially using LoRA for efficient parameter updates, significantly improves both the syntactic quality and execution accuracy of generated SPARQL queries.
Notably, our best-performing configuration, based on the LLaMA-3-8B-Instruct model fine-tuned for 15 epochs, achieves state-of-the-art results across all evaluation metrics, including BLEU, ROUGE, and relaxed exact match (RelaxedEM).
While RAG does not enhance performance in the presence of fine-tuning, this points to the importance of high-quality context retrieval.
FIRESPARQL offers a reproducible and configurable framework that can be adapted based on resource availability, paving the way for more robust, interpretable, and scalable QA systems over SKGs.


%
%
%
\bibliographystyle{splncs04}

%

\end{document}